\documentclass[%
 reprint,
 amsmath,amssymb,
 aps,
]{revtex4-2}

\usepackage{graphicx}% Include figure files
\usepackage{dcolumn}% Align table columns on decimal point
\usepackage{bm}% bold math
\usepackage{framed,multirow}
\usepackage{booktabs}
\usepackage{threeparttable}
\usepackage{amssymb}
\usepackage{latexsym}
\usepackage{booktabs}
\usepackage{amstext}
\usepackage{amsmath}
\usepackage{url}
\usepackage{xcolor}
\usepackage{ulem}
\usepackage{hyperref}
\def\ourmodel{BROW}

\begin{document}

\preprint{APS/123-QED}

\title{BROW: Better featuRes fOr Whole slide image based on self-distillation}% Force line breaks with \\
\author{
  Yuanfeng Wu, Shaojie Li, Zhiqiang Du, Wentao Zhu \\
  Zhejiang Lab, Hangzhou, China \\
  \texttt{\{wuyf, lishaojie, zhiqiangdu, wentao.zhu\}@zhejianglab.com} \\
  %% examples of more authors
}
% \thanks{Corresponding author. }

%              %  but any date may be explicitly specified

\begin{abstract}
Whole slide image (WSI) processing is becoming part of the key components of standard clinical diagnosis for various diseases. However, the direct application of conventional image processing algorithms to WSI faces certain obstacles because of WSIs’ distinct property: the super-high resolution. The performance of most WSI-related tasks relies on the efficacy of the backbone which extracts WSI patch feature representations. Hence, we proposed \ourmodel, a foundation model for extracting better feature representations for WSIs, which can be conveniently adapted to downstream tasks without or with slight fine-tuning. The model takes transformer architecture, pretrained using self-distillation framework. To improve model's robustness, techniques such as patch shuffling have been employed. Additionally, the model leverages the unique properties of WSIs, utilizing WSI's multi-scale pyramid to incorporate an additional global view, thereby further enhancing its performance. We used both private and public data to make up a large pretraining dataset, containing more than 11000 slides, over 180M extracted patches, encompassing WSIs related to various organs and tissues. To assess the effectiveness of \ourmodel, we run a wide range of downstream tasks, including slide-level subtyping, patch-level classification and nuclei instance segmentation. The results confirmed the efficacy, robustness and good generalization ability of the proposed model. This substantiates its potential as foundation model for WSI feature extraction and highlights promising prospects for its application in WSI processing.
% \begin{description}
% Secondary publications and information retrieval purposes.
% You may use the \texttt{description} environment to structure your abstract;
% use the optional argument of the \verb+\item+ command to give the category of each item. 
% \end{description}
\end{abstract}

\maketitle

%\tableofcontents

\section{Introduction}
\label{intro}
Deep learning is a technique based on artificial neural networks, which has achieved great success in many fields in recent years. It has the ability to process data with large amount, various dimensions and multi modalities, and learn complex features and patterns from data automatically, without the need for manual feature design, thereby improving the performance of data processing. In the field of
medical imaging, there also have been many successful applications. \citep{primakov2022automated} proposed a fully automated pipeline to detect and segment tumors in non-small cell lung cancer (NSCLC) CT images based on deep learning and region growing algorithm. \citep{eyuboglu2021multi} presented a multi-task weakly supervised learning approach for anomaly detection in whole-body FDG-PET/CT images. \citep{yu2020lymph} designed a transfer learning approach for prediction of lymph node metastasis in papillary thyroid carcinoma by leveraging radiomics features from other medical datasets.

These deep learning based approaches are making impressive progress in numerous tasks, nevertheless, achieved limited success in whole slide image (WSI) analysis.

With the emergence of digital pathology, WSI is becoming part of the key components of routine procedure for clinical diagnosis of many diseases. Traditional image processing algorithms face a number of challenges in this area. First, WSIs typically exhibit extremely high resolution to depict the details of cells and tissues. The substantial computational demands arising from high-resolution render the direct use of traditional convolution methods impractical. Many recent works are struggling with the trade-off between accuracy and computational efficiency due to the difficulty in processing large-scale gigapixel WSIs. Some approaches have employed multiple-instance learning (MIL) to partition the entire WSI into smaller patches for processing, followed by subsequent sophistic aggregation. For example, a Transformer based MIL framework was proposed by \citep{transmil} to explore both morphological and spatial information between different instances when making aggregation operator. \citep{milce} developed a MIL-based method to jointly learn both instance- and bag-level embeddings for making the final prediction. These methods all rely on a proficient patch-level feature extractor as a foundation, indicating the significance of constructing a well-work extractor. But most of these works have to retrain a task-specific model from scratch or use model pretrained on natural image datasets. Second, the models with insufficient parameters may struggle to effectively handle the abundant details and complex structures present in pathology images. Each pathology image possesses its specific morphological features and presentation patterns. The processes of collection, staining, scanning, and others can introduce image variations due to factors such as lighting conditions and equipment differences. Conventional algorithms are often designed for specific tasks, lacking robustness and generalization ability. Recently, large model has emerged as a viable option for addressing these problems, garnering increasing attention.

%大模型的发展、优势、应用，与WSI结合的前景
After achieving the considerable success in natural language processing (NLP), large model is steadily gaining more attention and gradually expanding into broader domains such as images and videos. For instance, \citep{CLIP} proposed the CLIP framework, which achieved excellent performance by pretraining the model to predict which caption corresponds to which image. \citep{ALIGN} utilized a dataset of over one billion noisy image-text pairs to expand visual and visual-linguistic representation learning. Compared with task-specific small models, rich representation learning and better generalization are parts of the advantages of these large models, also the key points in WSI processing. These works inspired us to integrate large model with WSI analysis to provide better feature representations, which can be further used across a wide range of downstream tasks. On the other hand, with techniques like prompt engineering and few-shot learning, large models are conveniently deployable to downstream tasks. This enables the trained model to have a broader range of application possibilities, fostering advancements in the corresponding researches.

In this work, we proposed a self-supervised learning approach to train a large-scale model for WSI feature extraction. There are three important elements for training the large-scale model: appropriate architecture, large dataset and suitable training method. Here, we integrated the vision transformer architecture as the backbone to extract feature representations from WSI slides. As for training data, we collected over 180M patches from more than 11000 WSI slides as a large dataset with both public and private data. The dataset contains slides stained with different methods. The slides are related to a variety of tissues and organs, like kidney, lung, breast, and so on. As for training method, our approach adapted the self-distillation framework to WSI's special properties. To leverage the multi-scale input of WSI, we utilize its hierarchical structure as additional global view to provide information at different scales. By adding views of color-augmented and patch-shuffled images, we motivate the model to learn transform-invariant features, ensure better generalization ability. We also integrate masked image modeling (MIM) technique into training to enhance the semantic learning. The evaluation is conducted on three downstream tasks over 10 datasets with easy adaptation. The experiment results verified that the proposed model can be used as a foundation backbone to extract better feature representations for WSI used in many analysis tasks. The main contributions of this work are listed as follows:
\begin{enumerate}
\item We established a large-scale foundation model for extracting better feature representation of WSIs.
\item By leveraging WSI's properties, we integrated color augmentation, patch shuffling, masked image model (MIM) and multi-scale input to add extra views into self distillation framework.
\item For pretraining the model, both private and public data are collected to make up a large dataset, containing more than 11000 slides, encompassing images about various organs and tissues. 
\item Comprehensive downstream experiments, including slide-level subtyping task, patch-level classification task and nuclei segmentation task, are performed on more than 10 datasets in total. 
\item With easy adaptation, the downstream experiment results demonstrate the superiority and robustness of the proposed method, indicating its promising potential to be used as backbone for extracting WSI feature representations.
\end{enumerate}

The remainder of this paper is organized as follows. In Section \ref{sect:relatew}, a brief introduction of related works was made. In Section \ref{sect:method}, the methods were elaborately illustrated. The experiments' settings and results were shown in Section \ref{sect:exp}. The discussion and conclusion are drawn in Section \ref{sect:diss} and \ref{sect:concl}.

\section{Related Work}
\label{sect:relatew}
\subsection{Deep Learning}
Deep learning has been employed as one of the solutions in the domain of medical image processing and achieved remarkable breakthroughs and successes. \citep{ardila2019end} proposed an end-to-end lung cancer screening method based on 3D deep learning, using low-dose chest CT images for automatic detection and diagnosis. By extracting features and classifying predictions, the method can accurately identify lung lesions with high sensitivity and specificity. \citep{fries2019weakly} develops a weakly supervised deep learning model for aortic valve malformation classification from unlabeled cardiac MRI sequences. By using large-scale, imperfect training labels, the model outperforms traditional supervised learning methods in performance. Deep learning based methods also have made some success in WSI processing field. \citep{barker2016automatedbrain} utilized a coarse-to-fine analysis of the localized characteristics in pathology images for WSI analysis to overcome the problem caused by large image size and rich tissue information. The first step of the analysis includes the extraction of spatially localized features. \citep{wang2019rmdl} designed a recalibrated multi-instance deep learning method for whole slide gastric image classification. The method is based on a localization network to extract features of the discriminating patches. These works of WSI processing highly rely on the quality of extracted features, but still have to train a task-specific model from scratch or use the model pretrained on ImageNet dataset\citep{imagenetdataset}. This motivates us to pretrain a model with WSI dataset to provide a domain-specific foundation, which can be used as a better initialization extracting better WSI features.
\subsection{Large Model}
With the emergence of self-supervised learning (SSL) and vision transformer (ViT), studies have revealed that increasing the parameter size of models or scaling up the training dataset often leads to enhanced model capabilities in downstream tasks, and even makes emergent capabilities appear on many complex problems \citep{kaplan2020scaling}. In the field of computer vision, many works have been conducted. In order to expand the training dataset, ALIGN \citep{ALIGN} uses a dataset of more than one billion noisy image-text pairs to expand visual and language representation learning, without taking complex data filtering and post-processing to clean the data. This study shows that downstream tasks can benefit from large dataset. \citep{DALLE} proposed a large-scale model, DALL·E, which is a generative model capable of generating corresponding images based on given textual descriptions. It utilizes the generative pretrained transformer architecture and self-attention mechanism, allowing it to capture contextual information from textual descriptions and translate it into guidance for image generation. Recently, \citep{segany} proposed SAM, made great progress in the field of natural images segmentation. It develops a large-scale prompt-based image segmentation model pretrained using a large dataset and achieves progress on a wide range of segmentation tasks. Overall, a growing body of research demonstrates the advantages of large-scale model for computer vision tasks. Inspired by these works, we expand the model and dataset to build a well-pretrained large model for better WSI feature representation learning.

\subsection{Self-supervised Training}

%intro of some well-known ssl paradigms
Constrained by insufficient annotated WSI data, self-supervised learning (SSL) is a feasible paradigm in visual representation learning. SSL is a deep learning method that does not require manual labeling of data. It extracts meaningful feature representations by learning the relationships and patterns among data itself without label information. There are two mainly used self-supervised representation learning techniques, contrastive learning and generative learning. The classical working Bert \citep{devlin2018bert} and GPT series \citep{radford2018gpt} \citep{radford2019gpt2} \citep{brown2020gpt3} in NLP are typical examples of using generative pretraining. Bert uses a random mask word and generates missing words in the training. The GPT series, on the other hand, use generation to predict what the next token will be for pretraining. Recently, generative learning has also been developed in the computer vision area, like MAE \citep{he2022masked} and SimMIM \citep{xie2022simmim} introduced the ViT architecture and proposed a MIM-based training method. These studies make it possible to train large models of computer vision. Compared with the generative learning approach, contrastive learning is more dominant for computer vision tasks. Specific frameworks for contrastive learning proposed by simCLR \citep{simclr} and MoCo \citep{moco} \citep{mocov2}, make large-scale pretraining possible by avoiding feature collapse with a large number of asymmetric positive and negative sample pairs. They require a large number of negative sample pairs during the training process, making training difficult. BYOL \citep{grill2020bootbyol} has simplified their training process by removing the memory bank. Later, SimSiam \citep{chen2021simsiam} removes the momentum encoder on top of that, making training more convenient. Inspired by these studies, DINO \citep{dino} combines the advantages of the ViT architecture with contrastive learning and self-distillation to achieve excellent performance in representation learning, obtained great performance in downstream classification and segmentation tasks. We adopt DINO as the base framework due to it's convenient to scale up the model and there is no need to construct the positive-negative pairs.  
% \section{Methods}
\section{Methods}
\label{sect:method}
\begin{figure*}[!t]
\centering
\includegraphics[width=\textwidth]{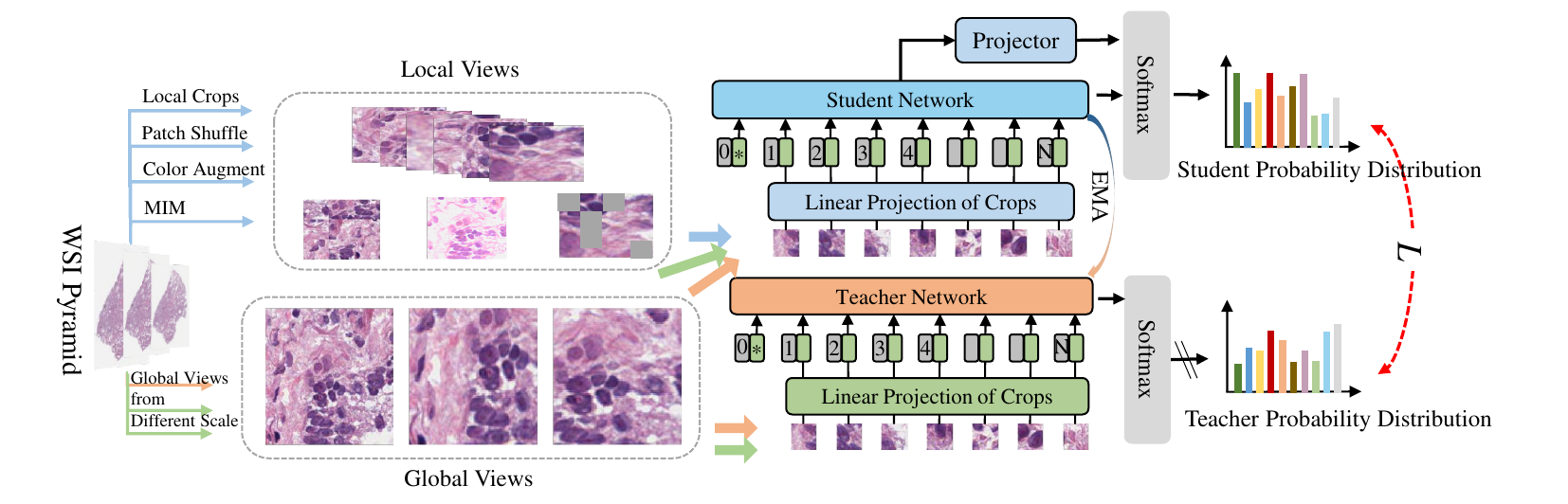}
\caption{The overall network framework. From the multi-scale pyramid of WSI, local views and global views were generated. Both global views and local views will be fed into student network, while only the global ones will be processed with teacher network. The probability distributions will be normalized with the softmax function to calculate the loss. The student network will be updated with gradients, while teacher network is updated using exponential moving average (EMA) on the student weights.}
\label{fig:network}
\end{figure*}

We adopt the self-distillation framework to learn feature representations. In order to address the inherent problems of WSI and leverage the distinct properties, we designed some methods, including color augmentation, masked image modeling (MIM), patch shuffling and utilizing the multi-scale input, to improve the robustness and quality of the feature representations. The overall network framework is shown in Fig.\ref{fig:network}.

\textbf{Base Framework} For pretraining, we adopt the self distillation paradigm, which trains a student network $g_{\theta_s}$ to match the output of a teacher network $g_{\theta_t}$. For self-supervised training of WSIs, first we construct a set $V$ of different views, containing images extracted from the same source image, using different distorted methods and crop strategies. This set consists of two parts: the local views and the global views. The whole set of views will be passed through the student network, getting the probability distribution $P_s(x)$ after normalized by a softmax function, while only the global views will be put into the teacher network, getting the distribution $P_t(x)$. Specifically, given an input view $x$, the probability distribution of the student network output is calculated with: 
\begin{equation}
    P_s(x)^{i} = \frac{\exp(g_{\theta_s}(x) ^{(i)} / \tau_s)}{ \sum ^K_{k = 1} \exp(g_{\theta_s}(x) ^ {(k)} / \tau_s)},\quad i= 1,2,...,K,
\end{equation}
where $K$ is the dimension of the model output, $\tau_s$ denotes a temperature parameter that controls the sharpness of the output distribution. The teacher network has a similar formula holds for $P_t$ with temperature $\tau_t$. To utilize WSIs' pyramid structure and improve the performance of dealing with images with different scales, we also use the corresponding multi-scale input as the additional global view. Moreover, in addition to the original approaches to get the views, we use color augmentation, MIM and patch shuffling to obtain the supplementary local views, in order to enhance the robustness of model. The main loss here is:
% \begin{equation}
%     L_m=\sum_{x\in\{x_1^g, x_2^g, x_3^g\}}\sum_{\stackrel{x'\in V}{x'\neq x}}H(P_t(x), P_s(x')),
% \end{equation}
% where $H(a,b)=-alogb$.

\begin{equation}
    L_m=\sum_{x\in\{x_1^g, x_2^g, x_3^g\}}\sum_{\substack{x'\in V \\ x'\neq x}}H(P_t(x), P_s(x')),
\end{equation}
where $H(a,b)=-alogb$.

\textbf{Color Augmentation} To ensure the robustness of the model facing data with color variants, including data from multi-centers or some processed with different staining methods, we use extra color augmentation to generate additional local view, driving the model to focus more on learning color-invariant representations. The augmentations includes changing the bright, saturation, color space of images within the batch, and so on. Given the color augmented image $x_c$, the loss $L_c$ of this view can be calculated as same as the main loss:
\begin{equation}
    L_{c} = H(P_t(x), P_s(x_c)),
\end{equation}

\textbf{MIM} Many works have proved that reconstructing the masked patches is a meaningful self-supervised task which drives the network to learn the latent representation of the image, such as MAE \citep{he2022masked} and SimMIM \citep{xie2022simmim}. We make a random mask $m$ for one input view of the student model and not do this for the teacher model, specifically, $x_m = m \odot x$. 

Instead of calculating the difference between the prediction and the unmasked patch at the image level, we add a cross-entropy loss between the output embeddings of both networks. Hence, the loss $L_{mim}$ has the same computation methodology with $L_m$, specifically,
\begin{equation}
    L_{mim} = H(P_t(x), P_s(x_m)),
\end{equation}
which can be conveniently integrated.

\textbf{Patching Shuffling} \citep{pirl} proposed PIRL to learn representations
that are invariant to the transformation and retain semantic information by using 
a commonly used pretext task, solving jigsaw puzzles. \citep{impash} introduced a similar approach to contrastive learning framework, learned invariant
representation while shuffling the input patches and improved the semantic quality. There are many trivial factors bringing variants to WSI. Inspired by these works, we use patch shuffling to generate another local view to learn the transform-invariant representations for improving the semantic quality and robustness. To encourage the representation of the shuffled image to be similar with that of its original counterpart, we use cross entropy loss $L_{ps}$ which can be defined as:
\begin{equation}
    L_{ps}=H(\text{Softmax}(e), \text{Softmax}(f(e_t))),
\end{equation}
where $f(\cdot)$ represents a multi-layer perceptron projector, $\text{Softmax}$ represents the softmax function, $e$ and $e_t$ denote the feature embeddings of the original and shuffled images, respectively. The whole loss can be defined as:
\begin{equation}
    L=L_m+L_c+L_{mim}+L_{ps}.
\end{equation}
The student network is updated with calculated gradients, while the teacher network is updated using the exponential moving average (EMA) paradigm on the student weights, specifically,
\begin{equation}
    \theta_t \leftarrow \lambda\theta_t + (1 - \lambda) \theta_s,
\end{equation}
with $\lambda$ following a cosine schedule from 0.996 to 1 during training. 

\section{Experiments and Results}
\label{sect:exp}
\subsection{Pretraining}
% du: 训练数据：统计中~  wsi 级 tcga_rcc: 3269  tcga_lung :3220, 'ShangtangData2': 1295, 'ShangtangData1': 1263,'ShangtangData3': 1159, 'Camelyon17': 1000  总patch 1亿8千万  180M

\subsubsection{Pretraining Datasets} We used both private and public data to make up the large pretraining dataset, including two parts from The Cancer Genome Atlas (TCGA) dataset \citep{tcga} (TCGA-RCC and TCGA-NSCLC), Camelyon 17 \citep{CAMELYON17} and three private datasets. There are more than 11000 slides of WSI. Each slide were cropped to patches with size of 256$\times$256 for training without using the label information. There are over 180 million patches extracted. Details can be found in Table\ref{pretrainingdatasets}.

\begin{table}[!t]
\caption{Datasets for pretraining. H\&E denotes Hematoxylin and Eosin Stain, IHC denotes Immunohistochemistry Stain, Pap denotes Papanicolaou Stain.}
\label{pretrainingdatasets}
\resizebox{\columnwidth}{!}{
\begin{tabular}{cccc}
\hline
Dataset & Slides & Organs & Stain \\
\hline
TCGA-RCC  & 3269 & Kidney & H\&E \\
TCGA-NSCLC  & 3220 & Lung & H\&E \\
Camelyon17  & 1000 & Breast & H\&E \\
Private Data1  & 1263 & Cervix & Pap \\
Private Data2  & 1295 & Breast & IHC \\
Private Data3  & 1159 & Digestive Tract & H\&E\\
\hline
\end{tabular}
}
\end{table}
\subsubsection{Pretraining Setting} The pretraining were performed on NVIDIA A100 GPUs using distributed mode. The code was developed based on Pytorch framework. The batch size was 1024, base learning rate was 0.0005, decayed with a cosine schedule. The models were trained for 100 epochs with 10 warm up epochs.

\subsection{Downstream Tasks}

\subsubsection{Downstream Datasets}

\begin{table}[!b]
\caption{Datasets for downstream tasks.}
\label{table:downdatasets}
\resizebox{\columnwidth}{!}{
\begin{tabular}{ccccc}
\hline
Dataset & Classes & Slides/Patches & Organs & Task \\
\hline 
CAMELYON16 & 2 & 399 & Breast & Slide-sub \\
TCGA-NSCLC & 2 & 1000 & Lung & Slide-sub \\
PANDA & 2 & 10,615 & Prostate & Slide-sub \\
TCGA-RCC & 3 & 896 & Kidney & Slide-sub \\
TCGA-BRCA & 2 & 973 & Breast & Slide-sub \\
% MHIST & 2 & 3,152 & Colorectal & Patch-cls \\
SIPaKMeD & 5 & 4,049 & Cervix & Patch-cls \\
Herlev & 2 & 917 & Cervix & Patch-cls \\
CoNSeP & / & 41 & Colon & Nuclei-seg \\
TNBC & / & 50 & Breast & Nuclei-seg \\
Kumar & / & 14 & Multi & Nuclei-seg \\
Lizard & / & 238 & Colon & Nuclei-seg \\
\hline
\end{tabular}
}
\end{table}
To evaluate the downstream task performance, we validate the pretrained models under three tasks: slide-level subtyping task, patch-level classification task and nuclei instance segmentation task. For slide-level subtyping task, we use CAMELYON16 \citep{bejnordi2017diagnostic}, TCGA-NSCLC, PANDA \citep{bulten2022artificial}, TCGA-RCC and TCGA-BRCA datasets. For patch-level image classification task, we use SIPaKMeD \citep{plissiti2018sipakmed} and Herlev \citep{jantzen2005pap} datasets. For nuclei instance segmentation task, we use CoNSeP \citep{graham2019hover}, TNBC \citep{tnbc}, Kumar \citep{kumar2017dataset} and Lizard  \citep{graham2021lizard} datasets. A brief summary is posted in Table\ref{table:downdatasets}. 

%可以放在supplementary material
CAMELYON16 is a public challenge dataset of sentinel lymph node biopsy of early-stage breast cancer, which consists of 399 H\&E stained WSIs (160 tumor, 239 normal). We split the training set by 9:1 into training and validation, respectively, then test on the official test set. The TCGA-NSCLC dataset includes two subtypes of lung adenocarcinoma (LUAD) and lung squamous cell carcinoma (LUSC). We randomly selected 1000 WSIs, including 500 LUAD and LUSC each, and divided them into training, validation and test sets with a ratio of 8:1:1 for cross-validation. The PANDA dataset is used for the classification of prostate cancer, consisting of 7,724 cancer samples and 2,891 non-cancer samples. They were split into training, validation and test sets according to the ratio of 7:1:2 for cross-validation. From TCGA-RCC, 896 WSIs were selected, of which the amounts of three subtypes were 90, 518 and 288. The distribution was kept unchanged and divided into training, validation, and test sets with a ratio of 8:1:1. The TCGA-BRCA dataset includes two subtypes: invasive ductal carcinoma (IDC) and invasive lobular carcinoma (ILC). 973 WSIs were selected, of which 774 IDCs and 199 ILCs were split into training, validation and testing sets, respectively, with a ratio of 8:1:1. SIPaKMeD includes 4,049 images of single cells with five classes, including Dyskeratotic, Koilocytotic, Metaplastic, Parabasal, Superficial-Intermediate, with 813, 825, 793, 787 and 831 samples, respectively. The Herlev dataset consists of 917 images of single cells. Within the dataset, there are 675 abnormal cells and 242 normal cells. Both SIPaKMeD and Herlev were split into train/val/test sets with a ratio of 3:1:1 for cross-validation. The segmentation model is trained on CoNSeP dataset, and tested with the official test splits. TNBC and Kumar were used for external tests. Here, we used the whole set of TNBC and the test set of Kumar split by \citep{graham2019hover}. The part of Lizard dataset with colon tissues was used as additional data discussed in Section\ref{sec:nucleiseg}. 

% TCGA-RCC is obtained by downloading and processing the tcga data. The specific steps are as follows: First, through the official download, a total of 3272 WSIs of 537 cases, 291 cases, and 113 cases are obtained, and then the "Sampe Type" is selected as "Primary Tumor". After wsi, through the pathological diagnosis data, filter the "primary diagnosis" as one of "Papillary adenocarcinoma", "Clear cell adenocarcinoma", "Renal cell carcinoma, chromosome type" data, and finally get the above results.The acquisition of TCGA-BRCA is similar to the above process, the difference is that "primary diagnosis" is selected as one of "Infiltrating duct carcinoma", "Lobular carcinoma"

\subsubsection{Downstream Experiments Setting} For slide-level subtyping task, we adopted the MIL framework with two advanced aggregators: CLAM by \citep{clam} and DTFD by \citep{zhang2022dtfdmil}. For
nuclei instance segmentation, we used the structure of Hover-Net \citep{graham2019hover} to separate the nuclei from the background. When performing cross-validation, the final results were calculated by averaging the metrics across all folds. Hyper-parameters are determined by the results on the validation set, while the performance was evaluated on the test set.

\subsubsection{Evaluation metrics}
For both slide-level subtyping task and patch-level classification task, we employed accuracy (ACC) and area under the curve (AUC) as evaluation metrics. For nuclei instance segmentation task, we adopted five metrics as \citep{graham2019hover}: DICE, Aggregated Jaccard Index (AJI) \citep{kumar2017dataset}, Detection Quality (DQ), Segmentation Quality (SQ) and Panoptic Quality (PQ). DICE is defined as:
\begin{equation}
    \text{DICE} = 2\times(X\cap Y)/(|X|+|Y|),
\end{equation}
where $X$ denotes the ground truth and $Y$ is the prediction. AJI computes the ratio of an aggregated intersection cardinality and an aggregated union cardinality between X and Y. PQ is defined as:
\begin{equation}
    \text{PQ}=\underbrace{\frac{|TP|}{|TP|+\frac{1}{2}|FP|+\frac{1}{2}|FN|}}_{DQ}\times \underbrace{\frac{\sum_{(x,y)\in TP}IoU(x,y)}{|TP|}}_{SQ}, 
\end{equation}
where x denotes a groundtruth (GT) segment, y denotes a prediction segment and IoU denotes intersection over union. TP, FN and FP denote matched pairs, unmatched GT segments and unmatched prediction segments, respectively.

\subsection{Results}

\begin{table*}[!t]
% \centering
\caption{ Results of slide-level multiclass subtyping task. The underline indicates the best result using the same framework, while the best results of two frameworks among all models were emphasized in bold.}
\label{table:slidecls}
\resizebox{\textwidth}{!}
{
\begin{tabular}{lccccccccccc}
\hline
\multirow{2}{*}{} & \multirow{2}{*}{} & \multicolumn{2}{c}{CAMELYON16} & \multicolumn{2}{c}{TCGA-NSCLC} & \multicolumn{2}{c}{PANDA} & \multicolumn{2}{c}{TCGA-RCC} & \multicolumn{2}{c}{TCGA-BRCA} \\ \cmidrule(lr){3-4} \cmidrule(lr){5-6} \cmidrule(lr){7-8} \cmidrule(lr){9-10} \cmidrule(lr){11-12}
\multirow{2}{*}{} & \multirow{2}{*}{} & ACC & AUC & ACC & AUC & ACC & AUC & ACC & AUC & ACC & AUC \\
\hline
\multirow{2}{*}{DINO\_Res50\_ImgNets} & CLAM & 0.8248 & 0.8408 & 0.8060 & 0.9058 & 0.8974 & 0.9632 & 0.8911 & 0.9657 & 0.7856 & 0.8121 \\
                                      & DTFD & 0.8574 & 0.8609 & 0.8244 & 0.8947 & 0.9036 & 0.9557 & 0.9044 & 0.9674 & 0.7927 & 0.8175 \\
\hline
\multirow{2}{*}{DINO\_Res50\_WSIs} & CLAM & 0.8574 & 0.9381 & 0.8365 & 0.9199 & 0.8984 & 0.9662 & 0.9100 & 0.9828 & 0.8021 & 0.8352 \\
                                   & DTFD & 0.8744 & 0.9522 & 0.8470 & 0.9222 & 0.9115 & 0.9629 & 0.8956 & 0.9788 & 0.8134 & 0.7689 \\
\hline
\multirow{2}{*}{DINO\_ViT-b\_ImgNets} & CLAM & 0.9000 & 0.9402 & 0.8500 & 0.9325 & 0.9176 & 0.9682 & 0.9256 & 0.9860 & 0.8577 & 0.8838 \\ 
                                      & DTFD & 0.9078 & 0.9244 & 0.8719 & 0.9433 & 0.9264 & 0.9718 & 0.9144 & 0.9792 & 0.8485 & 0.8747 \\
\hline
\multirow{2}{*}{DINO\_ViT-b\_WSIs} & CLAM & 0.9109 & 0.9662 & 0.8726 & 0.9488 & 0.9315 & 0.9763 & 0.9422 & 0.9895 & 0.8639 & 0.8962 \\ 
                                   & DTFD & 0.9124 & 0.9730 & 0.8889 & 0.9476 & 0.9372 & 0.9770 & 0.9556 & 0.9903 & 0.8773 & 0.8933 \\ 
\hline
\multirow{2}{*}{\ourmodel\_ViT-b\_WSIs} & CLAM & \uline{\textbf{0.9535}} & \uline{0.9756} & \uline{0.8818} & \uline{\textbf{0.9606}} & \uline{0.9407} & \uline{0.9802} & \uline{0.9511} & \uline{\textbf{0.9942}} & \uline{\textbf{0.8897}} & \uline{\textbf{0.9224}} \\ 
                                        & DTFD & \uline{0.9419} & \uline{\textbf{0.9787}} & \uline{\textbf{0.9007}} & \uline{0.9574} & \uline{\textbf{0.9434}} & \uline{\textbf{0.9813}} & \uline{\textbf{0.9611}} & \uline{0.9917} & \uline{0.8804} & \uline{0.9148} \\ 
\hline
\end{tabular}
}
\end{table*}

\subsubsection{Slide-level Subtyping Task}
To demonstrate the significance of the data and model scale, we run experiments of slide-level subtyping tasks on 5 different datasets: CAMELYON16, TCGA-NSCLC, PANDA, TCGA-RCC and TCGA-BRCA, employed two popular MIL frameworks: CLAM and DTFD. In this task, the pretrained model can be used as the feature extractor directly without any further fine-tuning. The feature aggregators were trained with data of different domains: WSIs and natural images (ImageNet dataset) with small scale (ResNet50) and large scale (ViT-b). The results are displayed in Table\ref{table:slidecls}. 

\begin{figure}[!t]
% \centering
\includegraphics[scale=0.5]{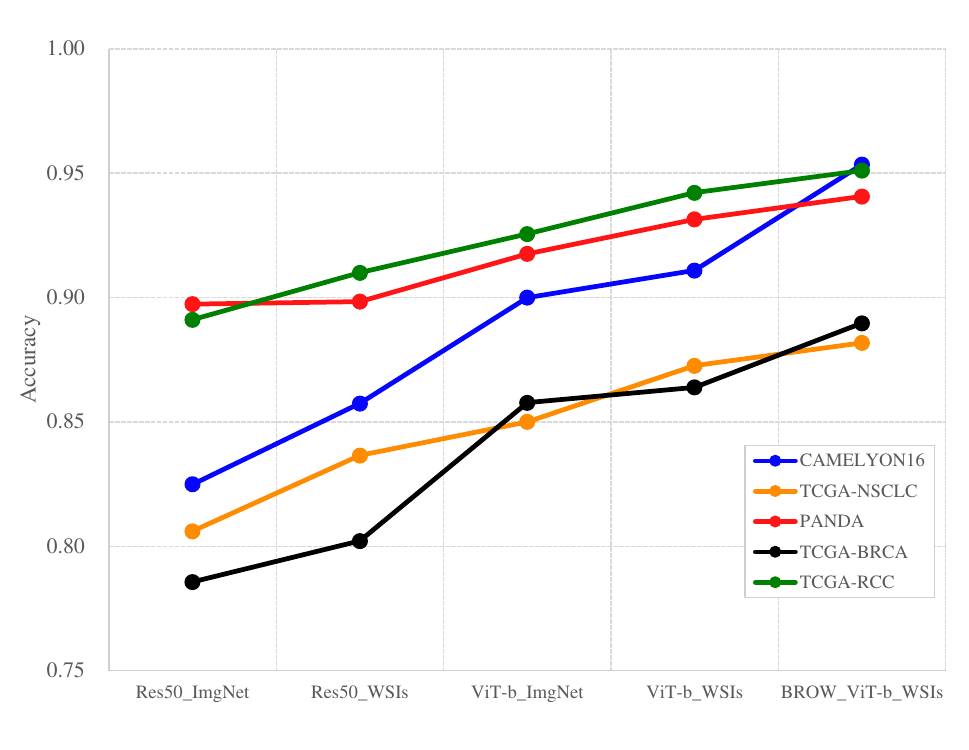}
\caption{The experiment results using different scale models pre-trained on different domains of data, taking accuracy as comparison metric.}
\label{fig:slidelineplot}
\end{figure}

By pretraining with data of specific domains, regardless of using Res50 or ViT-b as feature extractor, there has been significant improvement in subtyping performance. For Res50, when pretrained on WSIs instead of ImageNet, the model performance improved most on CAMELYON16 dataset, with +3.3\% of accuracy and +0.1 of AUC, using CLAM. For ViT-b, when pretrained on WSIs instead of ImageNet, the model performance improved most on TCGA-RCC dataset, with +4.1\% of accuracy and +0.01 of AUC, using DTFD. By pretraining with larger scale of model, regardless pretrained with ImageNet or WSIs, there has been prominent progress in subtyping performance. When pretrained with ImageNet, using ViT-b as feature extractor instead of Res50 yields improvements up to +7.5\% of accuracy and +0.1 of AUC on CAMELYON16 dataset using CLAM. When pretrained with WSIs, using ViT-b as feature extractor instead of Res50 leads to improvements up to +6.4\% of accuracy and +0.1 of AUC on TCGA-BRCA dataset using DTFD. The upward trend is conspicuously apparent in the line chart, as shown in Fig.\ref{fig:slidelineplot}. 

% When training was performed using WSI dataset instead of the ImageNet dataset, both ResNet50 and ViT-B models exhibited impressive performance improvements. 

The enhancements in performance caused by changing the pretraining datasets and expanding the model solidify the significance of training with domain specific data and the promising potential of applying larger model. It is worth pointing out that there are two outliers when use Res50 trained with WSIs as feature extractor on BRCA and RCC dataset. The decline of performance may
caused by the worse convergence limited by the model parameter scale. Compared with ImageNet dataset, the WSI dataset is more than ten times larger. The incongruity between the size of model and dataset leads to the underperformance. However, larger model, like ViT-b, did benefit from the large WSI dataset and further improve the subtyping performance, indicating the potential of using larger model as backbone. 

The bottom row of Table\ref{table:slidecls} is the results of the proposed model \ourmodel. By integrating the multiscale inputs and other methods, like MIM and color augmentation, the proposed model showed robust and competitive results on the five datasets, further enhancing the performance. 
% The underline indicates the best results using the same MIL framework, CLAM or DTFD. 

\textbf{Compared with Advanced Works}
There are many works trying to improve the performance of slide-level subtyping tasks. Most of them adopt the MIL framework and focus on exploring the correlation between instances and the whole slide. For instance, CLAM \citep{clam} used attention-based learning to identify valued sub-regions and uses instance-level clustering to constrain the feature space. Later, DTFD \citep{zhang2022dtfdmil} proposed a double-tier MIL framework to effectively use the intrinsic feature by introducing the concept of pseudo-bags. Some works also pay attention to the feature extractor, like SCL-WC \citep{scl-wc} proposed task-agnostic self-supervised feature pre-extraction and task-specific weakly-supervised feature refinement and aggregation for WSI-level prediction. Here, we made a comparison between \ourmodel~ and these advanced works, including CLAM, DTFD, TransMIL\citep{transmil}, ZOOMMIL\citep{zoommil}, DSMIL\citep{dsmil} and SCL-WC.

\begin{figure}[!t]
% \centering
\includegraphics[scale=0.5]{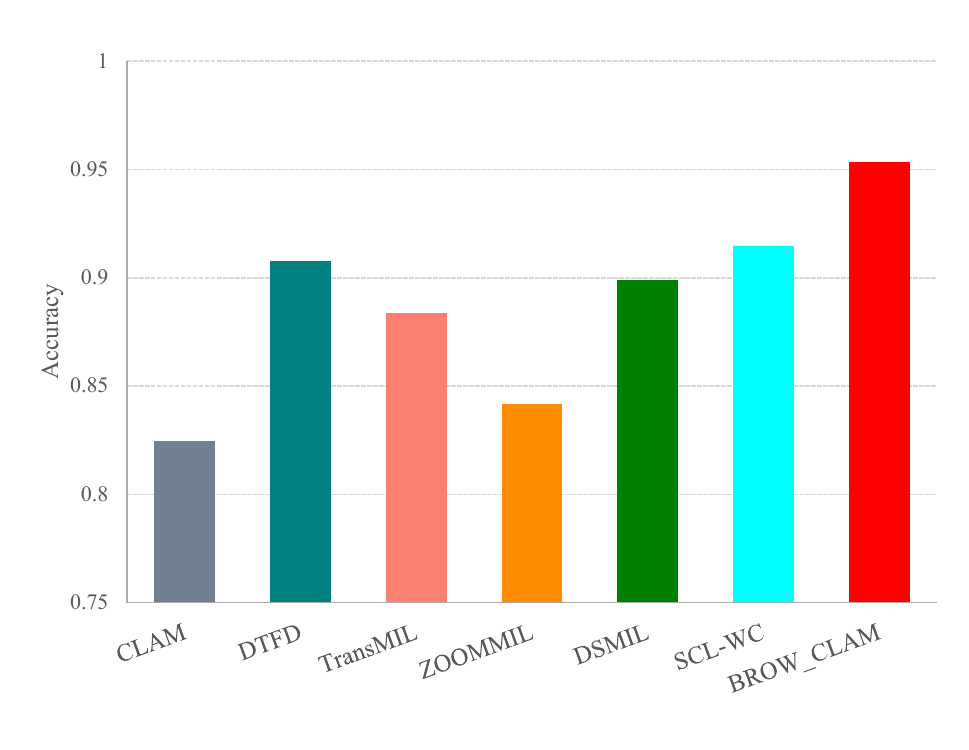}
\caption{The experiment results of advanced models on CAMELYON16.}
\label{fig:slidesota}
\end{figure}

Because Camelyon16 has the official set splits, we run experiments on this dataset. From the results shown in Fig.\ref{fig:slidesota}, by setting an exquisite aggregator or feature extractor, the models obtained good results. DTFD, DSMIL and SCL-WC all have got accuracy values near 0.9. However, by pretraining a large-scale model with large WSI dataset and directly using it as the feature extractor, \ourmodel~achieved the best performance with CLAM as an aggregator, the aggregator once with the worst performance. The results indicating slide-level subtyping task can benefit from the pretrained WSI foundation model as feature extractor a lot. Also, it can provide a better benchmark for developing aggregators in MIL framework. 

% \textbf{Visual results.}

% heatmap

\begin{table}[!b]
\vspace{-1.0cm}
% \renewcommand{\arraystretch}{1.2}
% \centering
\caption{Results of patch-level classification experiments. In fused model part, mix means the fusion of MobileNet, Inception V3 and Inception ResNet V2. Mix+\ourmodel~denotes introducing \ourmodel~into the fusion. Mix*+\ourmodel~ denotes using \ourmodel~ to replace the model performed worst in the former three. The fusion is realized based on fuzzy distance, details can be found at Section \ref{sec:patchcls}.}
\label{table:patchcls}
\resizebox{\columnwidth}{!}{
\begin{threeparttable}
\begin{tabular}{cccccc}
\hline
\multirow{2}{*}{} & \multirow{2}{*}{}  & \multicolumn{2}{c}{SIPaKMeD} & \multicolumn{2}{c}{Herlev} \\ 
\cmidrule(lr){3-4} \cmidrule(lr){5-6}
\multirow{2}{*}{} & \multirow{2}{*}{}  & ACC & AUC & ACC & AUC \\
\hline
\multirow{4}{*}{single model} & MobileNet  & 0.9585 & 0.9971 & 0.9476 & 0.9812 \\ 
\multirow{4}{*}{} & InceptionV3  & 0.9592 & 0.9970 & 0.9037 & 0.9658  \\
\multirow{4}{*}{} & Inception ResNet V2  & 0.9657 & 0.9984 & 0.9397 & 0.9842  \\
\multirow{4}{*}{} & \ourmodel    & \textbf{0.9783} & \textbf{0.9992} & \textbf{0.9663} & \textbf{0.9944}  \\
\hline
\multirow{2}{*}{fused model} & mix  & 0.9738 & / & 0.9488 & / \\
\multirow{2}{*}{} & mix+BROW  & 0.9793 & / & 0.9652 & /  \\
\multirow{2}{*}{} & mix*+BROW  & \textbf{0.9800} & / & \textbf{0.9718} & /  \\
\hline
\end{tabular}
% \begin{tablenotes}
% \footnotesize
% \item[1] mix represents the fusion of MobileNet, InceptionV3 and InceptionV2 models
% \end{tablenotes}
\end{threeparttable}
}
\end{table}

\begin{table}[!t]
\caption{Experiment results for data efficient adaptation. Adapter and adapter\_F are few-shot learning adapter with/without fine-tuning. LoRA represents the model fine-tuned using low-rank adaptation. Mix denotes combining the two terms. The number in bracket is the results trained with full data. The accuracy is used here as comparison metric.}
\label{cls_limitdata}
\resizebox{\columnwidth}{!}{

\begin{tabular}{lllllll}
\hline
           & \multicolumn{3}{c}{SIPaKMeD (97.83)}  & \multicolumn{3}{c}{Herlev (96.63)}    \\
\cmidrule(lr){2-4} \cmidrule(lr){5-7}
           
           & \multicolumn{3}{c}{using data ratio} & \multicolumn{3}{c}{using data ratio} \\
\hline
methods    & 1\%         & 5\%        & 10\%      & 1\%        & 5\%        & 10\%       \\
\hline
adapter    & 76.49       & 87.38      & 92.82     & 86.19      & 88.40      & 90.06      \\
adapter\_F & 76.73       & 90.10      & 93.81     & \textbf{86.74}      & 89.50      & 91.71      \\
LoRA       & 76.73       & 89.48      & 95.30     & 86.19      & 90.61      & 95.58      \\
mix        & \textbf{78.59} & \textbf{90.35} & \textbf{95.67} & \textbf{86.74} & \textbf{91.16} & \textbf{96.13}  \\
\hline

\end{tabular}
}
\end{table}

\subsubsection{Patch-level Classification}
\label{sec:patchcls}
We run experiments to test models' performance of patch-level classification task on two datasets, SIPaKMeD and Herlev. Here, in addition to adapting the deep-learning models to the task directly, we also use fusing method to further improve the classification performance. Previous works, \citep{bhowal2022fuzzy} and \citep{pramanik2022fuzzy}, have shown the benefits of model integration, which can effectively combine the strengths of individual models, resulting in higher accuracy and improved stability compared to a single model. We run experiments following the design mentioned in \citep{pramanik2022fuzzy}, employing four distinct models, including Inception V3 \citep{szegedy2016rethinking}, MobileNet V2 \citep{sandler2018mobilenetv2}, Inception ResNet V2 \citep{szegedy2017inception}, and the proposed \ourmodel. Confidence scores were extracted from trained models, and an ensemble method based on fuzzy distance was employed to aggregate these scores for final image category prediction.  

The results are shown in Table\ref{table:patchcls}. From the results we found that there are fluctuations of other  models' independent performance across two datasets. Inception V3 got competitive results on SIPaKMed, but underperformed others by a large margin on Herlev. However, \ourmodel~achieved the best and most robust results on both datasets, with accuracy of 0.9783 and 0.9663, respectively.  
By fusing the three models, Inception V3, MobileNet V2 and Inception ResNet V2, there are enhancements on both two datasets, demonstrating the benefits of model fusion. When integrating the proposed model \ourmodel~ into model fusion, the performance got further improved. By replacing the model in the fusion with the worst independent performance, the fused model obtained the best results, with accuracy of 0.9800 and 0.9718, respectively. 

\textbf{Data Efficient Deployment}
The proposed model is convenient to be adapted to this patch classification task with few shot learning. Inspired by the work of \citep{zhang2022tip}, we build the classification head by combining the advantages of parameter efficient fine-tuning and adapter which leverages a key-value cache model from the few shot training set. Given the support set with $K$ labeled images in each of $N$ categories, denoted as $I_K$, with their labels $L_N$, the feature representations of the images are extracted by \ourmodel~ first. The feature vector $\text{F}_{sup}$ and its corresponding label vector $\text{L}_{sup}$ can be defined as:
\begin{equation}
    \text{F}_{sup} = \text{\ourmodel}(I_K),\quad \text{F}_{sup}\in \mathbb{R}^{NK\times D},
\end{equation}
\begin{equation}
    \text{L}_{sup} = \text{OneHot}(L_N),\quad \text{L}_{sup}\in \mathbb{R}^{NK\times N},
\end{equation}
where $D$ is the dimension of extracted features. For the key-value cache, we treat $\text{F}_{sup}$ as keys, while the corresponding label vector $\text{L}_{sup}$ are their values. For a query image from test set, the feature is $f_{test}\in\mathbb{R}^{1\times D}$. The affinities $A$ between the query and keys can be estimated as:
\begin{equation}
    A=\text{exp}(-\beta(1-f_{test}\text{F}_{sup}^T)),\quad A\in\mathbb{R}^{1\times NK},
\end{equation}
where $\beta$ is a hyper-parameter for controlling the sharpness and $\text{exp}(\cdot)$ denotes the exponential function. The output logits of the test image by adapter are then caculated as $A\text{L}_{sup}$. The original method also fine-tuned the adapter via SGD to mitigate the data gap. We used method different from that. By parameter efficient fine-tuning methods, here we used LoRA \citep{hu2021lora}, the model can be adapted into the downstream task with limited data and computation. The new feature vector $\text{F}_{sup}^{'}$, $f_{test}^{'}$ and affinities $A^{'}$ can be calculated with the fine-tuned model in same way. Then we combine these two terms as final logits to capture the downstream task specific features while preserving the universal characteristics:
\begin{equation}
    \text{logits}=\alpha A\text{L}_{sup} + \alpha^{'} A^{'}\text{L}_{sup},
\end{equation}
where $\alpha$ and $\alpha^{'}$ are parameters to balance these two terms. 

The experiment results are shown in Table\ref{cls_limitdata}. Without any training, the model already got promising outcomes as shown in the top row. Training with data from 1\% to 10\%, the performance ascended rapidly. After fine-tuning with limited labeled data, the proposed model is able to obtain competitive results which is closed to the performance of model trained with all data. Especially for Herlev dataset, the model can beat other models' performance trained with all data with only 10\% data. What's more, combining the task specific feature with universal characteristics is beneficial for improving the performance, as shown in the bottom row the model got the best results.

\begin{figure}[!t]
% \centering
\includegraphics[width=\columnwidth, trim={0cm 0.2cm 0cm 0.5cm}]{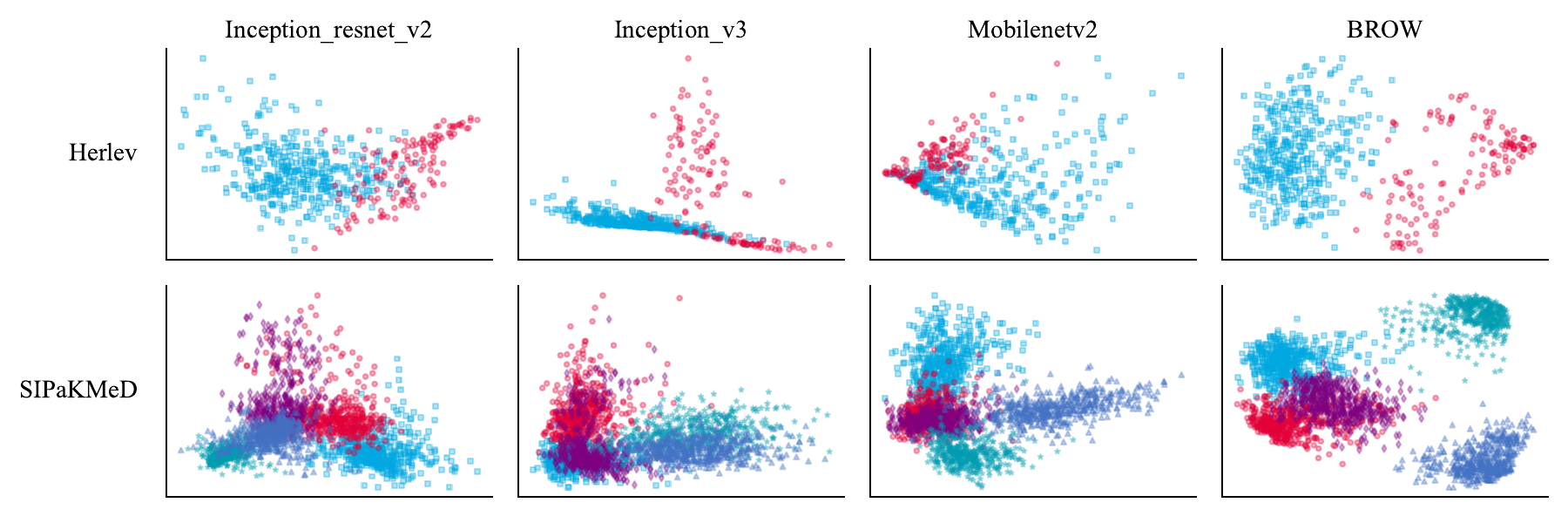}
\caption{The feature distribution provided by different models.}
\label{fig:patchscatter}
\end{figure}
\textbf{Visual Results}
The Fig.\ref{fig:patchscatter} displays the feature distribution provided by these models. In visualizing scatter plots of pre-extracted features, the proposed model provide best representation quality, on both two datasets. The  quality of representation is consistent with the final prediction performance.

The experiments demonstrate the promising and robust performance of the proposed model, both individually and as part of a hybrid ensemble. The proposed model can be easily adapted into patch-classification task with only limited labeled data and efficient adaptation.

\subsubsection{Nuclei Instance Segmentation}
\label{sec:nucleiseg}
Segmentation of cells and nuclei is a significant first step towards automatic analysis of WSI. Trained models are becoming important tools to assist pathologists. There have been several works that introduced SSL methods into WSI pretraining using vision transformer architectures to address various problems. We adapt some of the advanced works' trained models into this task. \citep{Scaling2022gigapixel} proposed a hierarchical image pyramid transformer (HIPT) using self-distillation framework. \citep{kang2023benchmarking} also provided a pretrained vit-based model using DINO, trying to establish a benchmark for SSL on pathology datasets. There is no abbreviation in the Paper. For simplicity, we call it B-DINO here. \citep{ctrans} proposed a hybrid model, called CTransPath, pretrained using MOCO V3 manner to learn universal feature representations for WSI. We adopted the Hover-Net \citep{graham2019hover} as framework for the nuclei segmentation task. There are two stages in this framework: in linear stage we freeze the parameters of feature extractors and only fine-tune the linear segment head; in fine-tune stage we fine-tune the whole model with downstream data. To integrate transformer into Hover-Net as the backbone, we use the feature map from the final layer of transformer to build a feature pyramid, using an approach validated as viable in the study conducted by \citep{li2022vitdet}. The main experiments were performed on the CoNSep dataset.

\begin{figure}[tphb]
\centering
\includegraphics[width=\columnwidth, trim={0cm 1.5cm 0cm 1.3cm}]{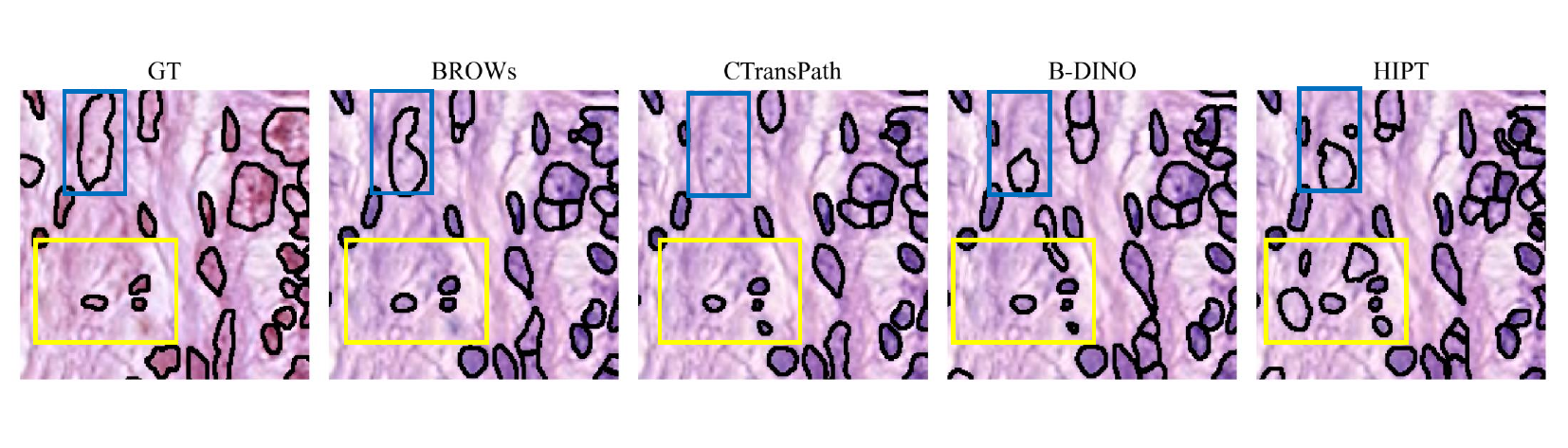}
\caption{Visual segmentation results. GT denotes the ground truth.}
\label{fig:segres}
% \vspace{-0.6cm}    
\end{figure}

\textbf{Compare with Advanced Works}
\begin{table}[]
\caption{Comparative experiment results of nuclei instance segmentation on CoNSeP dataset. \ourmodel s\_n represents fine-tuning with n shots of downstream data.}
\label{table:consep}
\resizebox{\columnwidth}{!}{\begin{tabular}{lllllll}
\hline
                          &                          & DICE   & AJI    & DQ     & SQ     & PQ     \\
\hline
\multirow{4}{*}{Linear}   & HIPT                     & 0.7789 & 0.4235 & 0.5138 & 0.7002 & 0.3617 \\
                          & B-DINO                   & 0.7773 & 0.4204 & 0.4905 & 0.6802 & 0.3349 \\
                          & CTransPath                & 0.7374 & 0.3384 & 0.4124 & 0.6664 & 0.2762 \\
                          & \ourmodel s               & \textbf{0.7862} & \textbf{0.4456} & \textbf{0.5426} & \textbf{0.7024} & \textbf{0.3825} \\
\hline
\multirow{4}{*}{Fine-tune} & HIPT                     & 0.7886 & 0.4273 & 0.5075 & 0.7030 & 0.3588 \\
                          & B-DINO                   & 0.7902 & 0.4352 & 0.5213 & 0.7085 & 0.3707 \\
                          & CTransPath                & 0.8061 & 0.4693 & 0.5677 & 0.7232 & 0.4119 \\
                          & \ourmodel s\_1     &0.6302  &0.1577  &0.2242   &0.6749  &0.1511             \\
                          & \ourmodel s\_2     &0.6837  &0.1736  &0.2554   &0.6685  &0.1708             \\
                          & \ourmodel s\_5     &0.7881  &0.4517  &0.5534   &0.7138  &0.3962   \\
                          & \ourmodel s                & \textbf{0.8122} & \textbf{0.4824} & \textbf{0.5870} & \textbf{0.7286} & \textbf{0.4292} \\
\hline
\end{tabular}}
\end{table}
To maintain consistency with the comparative objects, we use ViT-small as the backbone size to train our model, recorded as \ourmodel s. Table\ref{table:consep} demonstrates the quantitative results and Fig.\ref{fig:segres} displays the visual segmentation results. As shown in Table\ref{table:consep}, in both linear evaluation results and fine-tuning evaluation results, \ourmodel s achieved the best results. In linear test, HIPT, B-DINO and \ourmodel s outperform CTransPath by a large margin. Fine-tuning the net initialized with the pretrained weights, the performance of all the models improved, especially CTransPath. \ourmodel s still maintained better results compared to competing models over all the metrics. \ourmodel~ has the best ability to distinguish nuclear pixels from the background, verified by the high DICE score. CTransPath obtained sub-optimal results, suffering from low DQ values. That indicates its inability to detect the instance, leading to some overlook of detection. As shown in Fig.\ref{fig:segres}, CTransPath overlooks the large instance at the top left corner (noted by blue box). HIPT and B-DINO have low DICE and AJI score, penalized by the wrong-segmented area, indicating their bad segmentation performance. As shown in Fig.\ref{fig:segres}, they made wrong predication at the position noted by yellow box.

In Table\ref{table:consep}, we also list the results fine-tuned with limited data to evaluate the adaptation efficiency. Acquiring annotations with high-quality demands a significant investment of time and need high level of medical expertise. Data efficiency is important in this task. Fine-tunig with increasing amounts of data, the performance rapidly improved. The DICE score of \ourmodel s~ is close to HIPT fine-tuned with all data. Considering the PQ metric for both accurate quantification and interpretability to assess the performance, the proposed model can be efficiently adapted into this segment task with fine-tuning on five images, outperformed HIPT and B-DINO trained with all data. These experiments demonstrate that the proposed model can be efficiently adapted to this segment task with limited annotated data.

\textbf{Comparison with External Datasets}
To verify the generalization ability, we run experiments on two extra datasets: TNBC \citep{tnbc} and Kumar \citep{tnbc}. We adopted the models trained on CoNSeP dataset, and independently test them on TNBC and Kumar, in order to evaluate the ability for generalising to new organs and different sources. For Kumar, we used the test set split by Hover-Net \citep{graham2019hover}. For TNBC, we used the whole dataset as the test set. As shown in Table\ref{table:segexternaltest}, \ourmodel~generalized best on both two datasets. HIPT showed poor performance with unseen data, less effective at detecting nuclear pixels, which is reflected by the low DICE score. CTransPath and B-DINO achieved competitive performance with \ourmodel~on SQ score, indicating the close segmentation quality. However, SQ is calculated only within true positive segments and should therefore be observed together with DQ. Thus the overall segmentation performance of \ourmodel~is superior.

\begin{table}[!t]
\caption{Comparative experiment results on external test set.}
\label{table:segexternaltest}
\resizebox{\columnwidth}{!}{\begin{tabular}{lllllll}

\hline
\multirow{2}{*}{} & &\multicolumn{5}{l}{TNBC} \\
\hline
                  &                  & DICE   & AJI    & DQ     & SQ     & PQ          \\
\hline
                  & HIPT            &0.7546  &0.5460  &0.6845   &0.7428  &0.5121      \\
                  & B-DINO          &0.7668  &0.5736  &0.7290   &0.7463  &0.5476       \\
                  & CTransPath      &0.7610  &0.5801  &\textbf{0.7284}   &0.7443  &0.5468       \\
                  & \ourmodel s     &\textbf{0.7678}  &\textbf{0.5819}  &0.7283   &\textbf{0.7504}  &\textbf{0.5511}     \\
                  \hline
\multirow{2}{*}{} & &\multicolumn{5}{l}{Kumar} \\
\hline
                  &                   & DICE   & AJI    & DQ     & SQ     & PQ    \\
\hline
                  & HIPT             &0.7811  &0.4696  &0.5456   &0.7012 	&0.3874  \\
                  & B-DINO           &0.7706  &0.5094  &0.6153   &0.7098 	&0.4411   \\
                  & CTransPath       &0.7469  &0.4867  &0.5799   &0.7112 	&0.4159   \\
                  & \ourmodel s      &\textbf{0.7905}  &\textbf{0.5367}  &\textbf{0.6423}   &\textbf{0.7198} 	&\textbf{0.4653} \\
                  \hline
                  
\end{tabular}}
\end{table}

\begin{table}[!b]
\caption{Comparative experiments with models at different scales.}
\label{table:consepscale}
\resizebox{\columnwidth}{!}{\begin{tabular}{lllllll}
\hline
                          &                          & DICE   & AJI    & DQ     & SQ     & PQ     \\
\hline
\multirow{2}{*}{Linear}   & \ourmodel s               & 0.7862 & 0.4456 & 0.5426 & 0.7024 & 0.3825 \\
                          & \ourmodel b               & \textbf{0.7982}   & \textbf{0.4621}  & \textbf{0.5488}   & \textbf{0.7034}      & \textbf{0.3873} \\
\hline
\multirow{2}{*}{Fine-tune} & \ourmodel s              & \textbf{0.8122}   & \textbf{0.4824}  & \textbf{0.5870}   & \textbf{0.7286}      & \textbf{0.4292} \\
                          & \ourmodel b                & 0.8106 & 0.4766 & 0.5797 & 0.7268 & 0.4228 \\
\hline
\end{tabular}}
\end{table}

\begin{table}[!b]
\caption{Experiment results with different fine-tuning methods.}
\label{table:conseplora}
\resizebox{\columnwidth}{!}{
\begin{tabular}{llrrrrr}
\hline
            & Methods         & \multicolumn{1}{l}{DICE} & \multicolumn{1}{l}{AJI} & \multicolumn{1}{l}{DQ} & \multicolumn{1}{l}{SQ} & \multicolumn{1}{l}{PQ} \\
\hline
     & Full fine-tune   & 0.8106                   & 0.4766                  & 0.5797                 & 0.7268                 & 0.4228                 \\
    Fine-tune & freeze   blocks & 0.8153                   & 0.5031                  & 0.6163                 & \textbf{0.7362}        & 0.4550                 \\
            & LoRA            & 0.8181                   & \textbf{0.5104}         & 0.6200                 & 0.7343                 & 0.4568                 \\
            & Adapter         & \textbf{0.8183}          & 0.5085                  & \textbf{0.6224}        & 0.7327                 & \textbf{0.4575}  \\
\hline
            
\end{tabular}}
\end{table}

\begin{figure}[!t]
\centering
\includegraphics[scale=0.5]{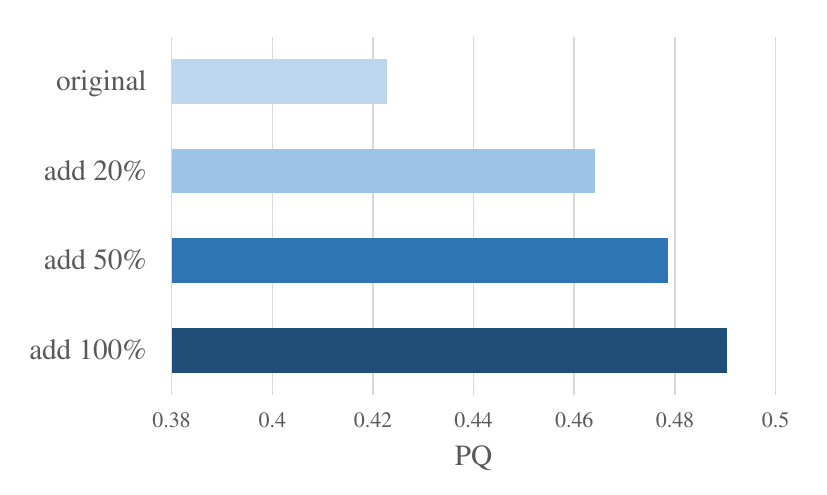}
\caption{Segmentation performance with additional data.}
\label{fig:segadddata}
\end{figure}

\textbf{Comparison with Model of Different Parameter Scale}
We run segmentation experiments using models trained with ViT-small and ViT-base architectures. The results are shown in Table\ref{table:consepscale}. According to the linear evaluation results, ViT-b based model obtained better results, indicating its better generalization performance. However, after fine-tuning the full parameters, the ViT-b based model was outperformed by ViT-s based model. One possible interpretation could be the over-fitting caused by the mismatching between limited data and large scale of parameters. To verify the hypothesis and address the over-fitting problem, we run experiments in two ways: using efficient fine-tuning methods and adding more data. To make a comparison among different fine-tuning methods, we adopted: 1) directly freezing part of the blocks; 2) using LoRA \citep{hu2021lora} to reduce the number of trainable parameters by injecting trainable rank decomposition matrices into each layer of the Transformer architecture; 3) using ViT adapter \citep{chen2022adaptformer} to efficiently transfer large pre-trained vision transformer models
to downstream tasks. We also post the full fine-tuning results as a baseline. The results are shown in Table\ref{table:conseplora}. The performance improved clearly, indicating the significance of fine-tuning methods in adapting a universal model into downstream tasks. By mitigating the over-fitting problem, our proposed model \ourmodel~achieved competitive results in nuclei instance segmentation task. To add more data into fine-tune training, we used the Lizard dataset \citep{graham2021lizard}. It's a large-scale dataset for colonic nuclear instance segmentation and classification. We leverage parts of the whole dataset from 3 different sources: GlaS, CRAG, and DigestPath. There are 238 patches in total of this additional dataset. We gradually add them into the fine-tune training and test the performance on the original test set. As shown in Fig.\ref{fig:segadddata}, when adding more data into the fine-tune training, even the data come from different sources, the model benefited from the extra data and improved the segmentation performance on the original dataset. 

\subsection{Ablation Study}
The nuclei segmentation experiment has both linear and fine-tune two stages, which is suitable for testing model's robustness and final performance. We run ablation studies on this task to test the proposed modules. The results are shown in Table\ref{table:ablation}. The linear stage and fine-tuning stage were performed on CoNSep dataset, while the external test were run on TNBC. 

From the results we found that in the linear stage, freezing the backbone and only fine-tuning the linear head, the proposed model achieved the best performance. Removing modules all leads to the decline of performance, which confirmed these modules can improve model's robustness. Removing the local view made by color augmentation leading to minimal decreasing of the results. This meets our expectations since we use colorjitter as base transform in other local views. This view used other color augmentation techniques to further increasing the robustness facing color variants. Removing the additional global view using multi-scale input and removing the MIM view result in the biggest performance decline. This indicates the multi-scale pyramid of WSI can provide valuable information and MIM can improve the quality of feature representations. In the fine-tuning stage, full fine-tuning all the parameters, the performance gap among models were reduced. This stage data influenced the performance more. But the model with all modules still obtained the best results. Then we run external test on TNBC dataset without any further fine-tuning. The model without MIM got the worst results, indicating MIM is significant in extracting the universal representation of images. 

% \usepackage{multirow}
\begin{table}[]
\caption{Experiments for exploring modules' influence on CoNSep and TNBC datasets. In the table, C denotes using color augments to add a local view. P means Patch shuffling. M denotes using MIM. G represents leveraging multi-sclale input to build extra global view.}
\label{table:ablation}
\resizebox{\columnwidth}{!}{
\begin{tabular}{llllllllll}
\hline
                               & C & P & M & G & DICE            & AJI             & DQ              & SQ              & PQ              \\
\hline
\multirow{5}{*}{Linear}        & \checkmark          & \checkmark  & \checkmark    & \checkmark            & \textbf{0.7513} & \textbf{0.3494} & \textbf{0.4408} & \textbf{0.6865} & \textbf{0.3042} \\
                               &         & \checkmark  & \checkmark    & \checkmark     & 0.7459          & 0.3383          & 0.4148          & 0.6777          & 0.2830          \\
                               & \checkmark          &   & \checkmark    & \checkmark     & 0.7393          & 0.3295          & 0.4239          & 0.6729          & 0.2863          \\
                               & \checkmark          & \checkmark  &     & \checkmark     & 0.7361          & 0.3305          & 0.3939          & 0.6701          & 0.2661          \\
                               & \checkmark          & \checkmark  & \checkmark    &      & 0.7334          & 0.3195          & 0.4047          & 0.6755          & 0.2752          \\
\hline
\multirow{5}{*}{Fine-tune}   & \checkmark          & \checkmark  & \checkmark    & \checkmark     & \textbf{0.7944} & \textbf{0.4355} & \textbf{0.5233} & 0.7090          & \textbf{0.3728} \\
                               &           & \checkmark  & \checkmark    & \checkmark     & 0.7930          & 0.4236          & 0.5026          & 0.7045          & 0.3560          \\
                               & \checkmark          &   & \checkmark    & \checkmark     & 0.7860          & 0.4314          & 0.5145          & 0.7009          & 0.3620          \\
                               & \checkmark          & \checkmark  &     & \checkmark     & 0.7928          & 0.4299          & 0.5064          & \textbf{0.7097} & 0.3611          \\
                               & \checkmark          & \checkmark  & \checkmark    &      & 0.7891          & 0.4307          & 0.5214          & 0.7076          & 0.3708          \\
\hline
\multirow{5}{*}{External Test} & \checkmark          & \checkmark  & \checkmark    & \checkmark     & \textbf{0.6791} & \textbf{0.4244} & \textbf{0.5492} & 0.6816          & \textbf{0.3765} \\
                               &          & \checkmark  & \checkmark    & \checkmark     & 0.6686          & 0.4160          & 0.5395          & 0.6747          & 0.3714          \\
                               & \checkmark          &   & \checkmark    & \checkmark     & 0.6721          & 0.3840          & 0.5186          & \textbf{0.6907} & 0.3601          \\
                               & \checkmark          & \checkmark  &     & \checkmark     & 0.6143          & 0.3869          & 0.4763          & 0.6259          & 0.3188          \\
                               & \checkmark          & \checkmark  & \checkmark    &     & 0.6632          & 0.4243          & 0.5286          & 0.6746          & 0.3606   \\
   \hline
\end{tabular}
}
\end{table}

\section{Discussion}
\label{sect:diss}
From the results of experiments and ablation study, the model achieved good performance on several downstream tasks, showing promising potential to be used as the backbone for extracting feature representations in WSI processing. The model can provide a better benchmark compared with models pre-trained on natural image datasets and improve the performance of tasks with good generalization ability and transferability, which has the potential to facilitate WSI processing related researches. By efficient adaptation, the foundation model can be easily deployed for different downstream tasks. Pretraining on a large dataset nurtures bountiful data diversity. The large-scale model provide the ability to process the big dataset and extract universal feature representations. By ablation studies, the additional local views building with color augmentations, MIM and patch shuffling enhance the robustness of the model. These methods drive the model to learn coloraugment-, mask- and transform-invariant features. With the leveraging of multi-scale input to build the extra global view, the model is able to better process WSI and have better generalization ability. The input pyramid introduces extra global view at different scale, providing various fields of view. The experiments in this study is diverse and comprehensive. There are three kinds of downstream tasks included, slide-level subtyping, patch-level classification and nuclei segmentation. These tasks are using data at different scales: patch level WSIs which can be processed directly with encoder and slide level WSIs which needs to be split first then processed with specific framework like MIL. In experiments, different downstream adaptation strategies are adopted. By freezing the parameter of the pretrained model or fine-tuning it with downstream tasks, the proposed model all achieved competitive results in corresponding tasks. To choose the adaptation strategy, first we should consider the task property. For example, the pretrained model can be used directly as the feature extractor in MIL framework when there is difficulty in fine-tuning the extacotr due to the large amount of patches of each slide. Second, the data amount also constrains the strategies. For small downstream datasets, it's better to use few-shot learning methods or parameter efficient learning approaches, such as LoRA. With these techniques, the pretrained model can be efficiently adapted to the downstream tasks with limited data and computation resources. 

However, during the experiments, there are also some bottlenecks restrict the research. Here, we list some of the factors which may facilitate the study:

\textbf{More data} There are more than 180M patches extracted from over 11000 slides, consisting of the large dataset. As mentioned before, the experiment results demonstrate the benefits of pre-training on a large dataset, as shown in Fig.\ref{fig:slidelineplot} and Table\ref{table:patchcls}. Then it comes a question: is the dataset large enough? Though the dataset has improved the models' performance in various downstream tasks, there still remains the necessity to expand the dataset. When compared with the well-known natural image dataset, ImageNet22k, which has about 14M images within the dataset, our dataset is quite large. However, some factors constrain the efficacy of the dataset. First, the  distribution of different staining techniques is unbalanced. Various staining methods are employed in WSI analysis for different purposes, like H\&E allows the observation of pathological alterations while the IHC enables the assessment of protein expression within the tissues. To make the pre-trained model become a better foundation model of WSI, we need to find more data using different staining methods, considering the unbalanced distribution now. Second, we need more data to enhance the class diversity. Still compared with ImageNet22k which has 22k classes of images, clinical data with WSI may not have so many classes. But enlarging the dataset's class diversity, like increasing data from different organs with various symptoms is a feasible approach to improve the generalization ability. Third, compared with natural images, WSIs manifest high similarities, which hinder the efficacy of large datasets. Hence, more data should be collected to construct a better `large' dataset.

\textbf{More Downstream Tasks.} Like ImageNet challenge for comparing models' performance in the natural image field, there are some well-known public challenges in WSI area, like Camelyon16, in which the proposed \ourmodel~got competitive results. Some datasets are wildly used, like TCGA-NSCLC, but there is no official selection of the data and the set splits for cross-validation. The setting will influence the performance a lot during reproduction. Some challenges are already addressed with pretty good results because of the factors such as the lack of diversity or the inherent intricacy of the task. More downstream tasks for WSI field as standard are needed, not only for providing more convincing results, but also can save researchers' time for reproducing prior works.

\textbf{More Flexible Model.} The multi-scale structure is the distinct property of WSI. Each scale contains a WSI image with different resolution.  We leverage this structure to construct extra global view. A more flexible model which can better use this property may further improve the efficiency and efficacy. For example, clinical experts can make fast detection using rough images and make further diagnosis using images with higher resolution, which can be integrated into auto-diagnosis pipeline to expedite the prediction. Task-specific models always focus on the maximum magnification. Some researchers have paid more attention to context details of this, like \citep{van2021hooknet} introduced concentric patches at multiple resolutions to high-resolution semantic segmentation. The pyramid structure provides the images at different resolutions, potentially being beneficial for the model to deal with data from different centers with various magnifications. Therefore, the model with more flexibility to deal with the multi-scale input is needed.

\section{Conclusion}
\label{sect:concl}
This study proposed a large-scale foundation model, called \ourmodel, for WSI processing. With the help of large dataset, scaled-up model size and appropriate training framework, the proposed model is able to extract better feature representations for WSI images. By directly being integrated to the original framework or efficiently adapting the model with downstream data, this model achieved competitive results on slide-level subtyping task, patch-level classification task and nuclei instance segmentation task. Through a comprehensive experiment comparison and analysis over ten datasets, the proposed model demonstrated its superiority, robustness and generalization ability.

% \nocite{*}

% \bibliography{apssamp}% Produces the bibliography via BibTeX.
\normalem
\bibliography{apssamp}
\bibliographystyle{unsrt} 

\end{document}